\documentclass{llncs}
\usepackage{theorem,enumitem,xspace,amssymb}
\usepackage{latexsym,prooftree,url}

\newcommand{\ov}[1]{\overline{#1}}

\newcommand{\lf}[1]{\mbox{\texttt{#1}}}

\newcommand{\ALC}{\mbox{${\cal ALC}$}\xspace}

\newcommand{\BCDL}{\mbox{\mbox{${\cal BC\!D\!L}$}}\xspace}
\newcommand{\BCDLZ}{\mbox{\mbox{${\cal BC\!D\!L}_0$}}\xspace}
\newcommand{\KALC}{\mbox{${\cal KALC}$}\xspace}

\newcommand{\NC}{\mbox{\tt NC}}

\newcommand{\NR}{\mbox{\tt NR}}
\newcommand{\NI}{\mbox{\tt NI}}
\newcommand{\VAR}{\mbox{\tt Var}}

\newcommand{\PAPdl}{\mbox{${\cal S}{\cal C}$}\xspace}

\newcommand{\serv}[1]{\mathsf{#1}}
\newcommand{\SERV}[4]{\serv{#1}(#2):: #3 \PAPrule #4}
\newcommand{\Impl}[1]{\Phi_{\mbox{\tiny #1}}}

\newcommand{\ity}{\mbox{\sc it}}
\newcommand{\true}{\mbox{\texttt{tt}}} 
\newcommand{\extract}[2]{\Phi^{#1}_{\Ncal}(#2)}

\newcommand{\stru}[1]{\langle #1 \rangle}

\renewcommand{\o}{\sqcup}
\newcommand{\e}{\sqcap}
\newcommand{\non}{\neg}
\newcommand{\im}{\!\sqsubseteq\!}

\newcommand{\seq}{\vdash}
\newcommand{\RuleLine}[2]{\rule{#1}{.7pt}~\mbox{#2}}

\newcommand{\real}[1]{\rhd\langle#1\rangle\,}

\newcommand{\dimo}[1]{
  \mbox{$\ \mid\hspace{-1.1 ex}
  \frac{\hspace{2 ex}}
       {\mbox{\hspace{0.6 ex}\tiny $#1$ \hspace{0.6ex}}}\ $}}

\newcommand{\dimBCDLZ}{\dimo{\BCDLZ}}

\newcommand{\Tab}[3]{ 
  \frac{\phantom{a}\stackrel{\textstyle #1} 
    {\phantom{\scriptscriptstyle.}}\phantom{a}} 
  {\stackrel{\phantom{\scriptscriptstyle .}}{\textstyle #2}} {\mbox{\tiny #3}} 
}

\newcommand{\Dcal}{{\cal D}}

\newcommand{\Lcal}{{\cal L}}
\newcommand{\Mcal}{{\cal M}}
\newcommand{\Ncal}{{\cal N}}

\renewcommand{\a}{\alpha}
\renewcommand{\b}{\beta}
\newcommand{\g}{\gamma}

\newcommand{\s}{\sigma}
\newcommand{\G}{\Gamma}

\newcommand{\subs}{\sqsubseteq}

\newcommand{\ND}[1]{{\cal N\!\cal D}_{{\rm #1}}} 
\newcommand{\CALC}{\ND{}}

\newcommand{\nome}[1]{\using{\scriptstyle #1}}

\newcommand{\Tbox}{\mbox{\sffamily T}_{PS}}

\newcommand{\EnvPS}{\mathbf{E}_{PS}}

\newcommand{\Tbo}{\mathbf{T}}

\newcommand{\Ln}{\Lcal_\Ncal}

\newcommand{\Env}{\mathbf{E}}
\newcommand{\pre}{\mbox{Pre}}
\newcommand{\post}{\mbox{Post}}
\newcommand{\PAPrule}{\Rightarrow}

\newcommand{\EndEs}{\mbox{}~\hfill$\Diamond$}

\title{A Note on Semantic Web Services\\ Specification and Composition in Constructive
  Description Logics\thanks{Part of this work will appear as a position paper 
  in Proceedings of the 4th International Conference on Web Reasoning and 
  Rule Systems (RR 2010).}}
\author{Loris Bozzato \and Mauro Ferrari}
\institute{Dipartimento di Informatica e Comunicazione\\
  Universit\`{a}  degli Studi dell'Insubria\\
  Via Mazzini 5, 21100, Varese, Italy}
\date{}

\begin{document}
\maketitle

\pagestyle{plain}

\vspace{-2ex}
\begin{abstract}
  The idea of the Semantic Web is to annotate Web content and services with
  computer interpretable descriptions with the aim to automatize many tasks
  currently performed by human users. In the context of Web services, one of the
  most interesting tasks is their composition. In this paper we formalize this
  problem in the framework of a constructive description logic. In particular we
  propose a declarative service specification language and a calculus for
  service composition. We show by means of an example how this calculus can be
  used to define composed Web services and we discuss the problem of automatic
  service synthesis.
\end{abstract}


\section{Introduction}

The idea of the Semantic Web is to annotate Web content and services with computer interpretable descriptions in order to automatize many tasks currently performed by human users. In the context of the Web services, this has led to the definition of \emph{semantic Web services}, that is a semantic description of the capabilities and the structure of services in the languages of the semantic Web.
The current proposals for the representation of semantic Web services, as
\mbox{OWL-S}~\cite{owls:2007}, view services as processes with pre- and post-
conditions and effects.  The representation by pre- and post- conditions
describe the requirements and output of a service that is useful to retrieve the
service; the representation of the process associated with a service describe
the interaction with other given services.
One of the main problems in the context of Web services is their composition.
The problem can be stated as follows: given a composition goal, represented as a
service with pre- and post- conditions, compose the available services so to
satisfy the goal.  Obviously in this context the challenge is to provide tools
to support the definition of the composite service or, at best, to automatize
the entire composition process. Using the well known relation between semantic
Web languages and description logics, here we discuss the problem of service
composition in the context of constructive description logics. This allows us to
draw from the long tradition of use of constructive mathematics in the context
of program synthesis.  Indeed, the composition calculus we discuss in
this paper is inspired by \cite{MigMosOrn:91}.

In this paper we formalize the composition problem in the framework of the
constructive description logic $\BCDLZ$. 
This paper represents an initial presentation for our approach:
its main contribution is to lay down the definitions for a composition
language in a way that it can then be possible to define an automatic procedure
for composition by software synthesis principles.
Moreover, our approach also exhibits an interesting application of constructive semantics
for description logics and demonstrates how to take advantage of their computational properties.

The logic $\BCDLZ$ that forms the base of our proposal is a subsystem of
$\BCDL$~\cite{FerFioFio:2010}, a logic based on an \emph{information terms
 semantics}. The main advantage of this semantics is to provide a natural
notion of state which is at the base of our formalization of Web services and
Web service composition. Moreover, how discussed in \cite{FerFioFio:2010} this
logic supports the proofs-as-programs paradigm. This allows to characterize in
this setting also the problem of automatic Web services composition.
For our purposes, in this paper we present a natural deduction calculus for $\BCDLZ$:
however, this logic can be related to $\KALC$~\cite{BozFerFioFio:10}, a constructive
description logic based on a Kripke-style semantics for which we provided
a decidable tableaux calculus.

In the following sections we introduce our formalism for the specification of
services and we present our calculus $\PAPdl$ for the definition of composite
services.  In order to do this, we begin by introducing the syntax and
information terms semantics of $\BCDLZ$.


\section{\BCDLZ: Syntax and Semantics}
\label{sec:logic}

$\BCDLZ$ is a subsystem of $\BCDL$ \cite{FerFioFio:2010} which is the
correspondent in the information terms semantics context of the basic
description logic $\ALC$~\cite{dlHandbook:03}.  The language $\Lcal$ for
$\BCDLZ$ is based on the following denumerable sets: the set $\NR$ of {\em role
  names}, the set $\NC$ of {\em concept names}, the set $\NI$ of {\em individual
  names} and the set $\VAR$ of {\em individual variables}.  The \emph{concepts}
$C,D$ and the
\emph{formulas} $K$ of $\Lcal$ are defined according to the following grammar:
\begin{eqnarray*}
  C,D &::=& A~|~\non C~|~C\sqcap D~|~C\sqcup D~|~\exists R.C~|~\forall R.C\\
  K &::=&\bot~|~(s,t):R~|~t:C~|~A\im C
\end{eqnarray*}
where $s,t\in\NI\cup\VAR$, $R\in\NR$, $A\in\NC$.  A \emph{closed formula} is a
formula not containing individual variables.  A \emph{simple formula} is either
a formula of the kind $\bot$, $(s,t):R$ or a formula of the kind $t:C$ with $C$
a concept name or a negated concept.  We remark that we do not allow general
inclusions of concepts, but we only admit atomic concepts in the antecedent of a
subsumption.

In the following we will be interested in the formulas generated by a finite
subset $\Ncal$ of $\NI$; we denote with $\Lcal_\Ncal$ such a language.  A {\em
  model} $\Mcal$ for $\Lcal_\Ncal$ is a pair
$(\Dcal^\Mcal,.^\Mcal)$, where $\Dcal^\Mcal$ is a non-empty set (the {\em
  domain} of $\Mcal$) and $.^\Mcal$ is a {\em valuation} map such that: (i) for
every $c\in\Ncal$, $c^\Mcal\in \Dcal^\Mcal$; (ii) for every $A\in\NC$, $A^\Mcal
\subseteq \Dcal^\Mcal$; (iii) for every $R\in\NR$, $R^\Mcal\subseteq
\Dcal^\Mcal\times \Dcal^\Mcal$.

A non atomic concept $C$ is interpreted by a subset $C^\Mcal$ of $\Dcal^\Mcal$
as usual: 
\[
\begin{array}{l}
  (\non C)^\Mcal \;\;=\;\; \Dcal^\Mcal\;\setminus\; C^\Mcal\\[1ex]
  (C\sqcap D)^\Mcal \;\;=\;\; C^\Mcal\cap D^\Mcal\\[1ex]
  (C\sqcup D)^\Mcal \;\;=\;\; C^\Mcal\cup D^\Mcal\\[1ex]
  (\exists R.C)^\Mcal \;\;=\;\; \{\,c\in\Dcal^\Mcal \;|\;
  \mbox{there is $d\in\Dcal^\Mcal$ s.t. $(c,d)\in R^\Mcal$ and $d\in C^\Mcal$}  \}
  \\[1ex]
  (\forall R.C)^\Mcal \;\;=\;\; \{\,c\in\Dcal^\Mcal \;|\;
  \mbox{for all $d\in\Dcal^\Mcal$, $(c,d)\in R^\Mcal$ implies $d\in C^\Mcal$}  \}
\end{array}
\]
A closed formula $K$ is {\em valid} in $\Mcal$, and we write $\Mcal \models K$,
if $K\neq\bot$ and: 
 \[
 \begin{array}{l}
   \Mcal \models (s,t):R~\mbox{iff}~ (s^\Mcal, t^\Mcal)\in R^\Mcal\\[1ex] 
   \Mcal \models t:C~\mbox{iff}~ t^\Mcal \in C^\Mcal\\[1ex]
   \Mcal \models A\,\im\, C\;\;\mbox{iff}\;\; A^\Mcal \subseteq C^\Mcal
 \end{array}
 \]
A theory $T$ consists of a \emph{TBox} and an \emph{ABox}.  A TBox is
a finite set of formulas of the form $A\im C$. An ABox is a finite set of
concept and role assertions: a \emph{concept assertion} is a formula of the kind
$c:A$, with $c\in\NI$ and $A\in\NC$; a \emph{role assertion} is a formula of the
kind $(c,d):R$, with $c,d\in\NI$ and $R\in\NR$.

The constructive interpretation of $\BCDLZ$ is based on the notion of
\emph{information term}~\cite{FerFioFio:2010}. 
Intuitively, an information term $\a$ for a closed formula $K$ is a
structured object that provides a justification for the validity of
$K$ in a classical model, in the spirit of the BHK interpretation
of logical connectives~\cite{Troelstra:99}.
Information terms are inductively defined on the structure of the closed formulas,
starting from the constant symbol $\true$ associated to atomic formulas.
The meaning and the correct reading of an information term is
provided by the related formula.
For instance, the truthness of an existential formula $c: \exists R.C$
in a classical model $\Mcal$ can be explained by its information term
$(d,\alpha)$, that explicitly provides the witness $d$ such that $(c^\Mcal, d^\Mcal)
\in R^\Mcal$ and $d^\Mcal \in C^\Mcal$; moreover, the information term $\alpha$
recursively explains why $d^\Mcal \in C^\Mcal$.

Formally, given $\Ncal\subseteq\NI$ and a closed formula $K$ of $\Lcal_\Ncal$,
we define the set of information terms $\ity_\Ncal(K)$ by induction on $K$ as
follows.
\[
\begin{array}{l}
  \ity_\Ncal(K)~=~\{\true\},\;
  \mbox{if $K$ is a simple formula}\\[1ex]
  \ity_\Ncal(c: C_1\sqcap C_2) \;= \;
  \{\, (\a,\b) ~|~ \mbox{$\a\in\ity_\Ncal(c: C_1)$ and $\b\in\ity_\Ncal(c: C_2)$ }\}\\[1ex]
  \ity_\Ncal(c: C_1\sqcup C_2) \;= \;
  \{\, (k,\a) ~|~ \mbox{$k\in\{1,2\}$ and $\a\in\ity_\Ncal(c: C_k)$ }\}\\[1ex] 
  \ity_\Ncal(c: \exists R.C) \;=\;
  \{\, (d,\a)  ~|~ \mbox{$d\in\Ncal$ and $\a\in\ity_\Ncal(d: C)$ }\}\\[1ex]
  \ity_\Ncal(c: \forall R.C) \;=\;  
  \{\, \phi :\Ncal\rightarrow \bigcup_{d\in\Ncal} \ity_\Ncal(d: C)
  ~|~ \phi(d)\in\ity_\Ncal(d: C)\,\}\\[1ex]
  \ity_\Ncal(A\im C) \;=\; 
  \{\, \phi :\Ncal\rightarrow \bigcup_{d\in \Ncal} \ity_\Ncal(d: C)
  ~|~ \phi(d)\in\ity_\Ncal(d: C)\,\}
\end{array}
\]
We remark that information terms for $K=c:\forall R.C$ and $K=A \im C$ formulas
are defined as a set of functions mapping every element $d$ of
the finite set $\Ncal$ to an information term for $d:C$. In other words,
any information term for these formulas justifies 
that every element of $\Ncal$ belongs to the concept defined by
$C$ in a given classical model.

Let $\Mcal$ be a model for $\Lcal_\Ncal$, $K$ a closed formula of $\Lcal_\Ncal$
and $\eta\in \ity_\Ncal(K)$.  We define the \emph{realizability relation}
$\Mcal\real{\eta} K$ by induction on the structure of $K$.
\begin{itemize}[leftmargin=*]
\item $\Mcal\real{\true} K$ iff $\Mcal\models K$.
\item $\Mcal\real{(\a,\b)}  c: C_1\sqcap C_2$ iff $\Mcal\real{\a} c: C_1$ and $\Mcal\real{\b}  c: C_2$.
\item $\Mcal\real{(k,\a)} c: C_1\sqcup C_2$ iff $\Mcal\real{\a} c: C_k$.
\item $\Mcal\real{(d,\a)} c: \exists R.C$ iff $\Mcal\models (c,d) : R$ and 
$\Mcal\real{\a} d: C$.
\item $\Mcal\real{\phi} c: \forall R.C$ iff $\Mcal\models  c: \forall R.C$ and, for every $d\in\Ncal$,
  $\Mcal\models (c,d):R$ implies $\Mcal\real{\phi(d)} d: C$
\item $\Mcal\real{\phi}A\im C$ iff, $\Mcal\models  A\im C$ and, for every $d\in\Ncal$,
    if $\Mcal\real{\true}d:A$ then  $\Mcal\real{\phi(d)} d: C$
\end{itemize}
If $\G$ is a finite set of closed formulas $\{K_1,\dots,K_n\}$ of $\Lcal_\Ncal$ 
(for any ordering of the formulas of $\G$),
$\ity_\Ncal(\G)$ denotes the set of $n$-tuples $\ov{\eta}=
(\eta_1,\dots,\eta_n)$ such that, for every $1\leq j\leq n$,
$\eta_j\in\ity_\Ncal(K_j)$; $\Mcal\real{\ov\eta}\G$ iff, for every $1\leq j\leq
n$, $\Mcal\real{\eta_j}K_j$.

Now, we introduce the example we refer to throughout this paper.
\begin{example}[Theory definition]
  Our example represents a reinterpretation and a formalization in our context
  of the ``purchase and delivery service'' example of~\cite{TraversoP:04}.  The
  example presents a system composed by three agents: a User, a Shipper and a
  Producer agent.  The Shipper and the Producer provide the User with services
  to request and obtain offers for the delivery and the purchase of a product:
  the goal of the example is to combine the services of the two agents in order
  to provide the User with a single service to request the production and
  shipping of a product.  We begin by defining the theory $\Tbox$ that models
  our system.\\[-3ex]
  \begin{center}\small
  $
  \begin{array}{l}
      \lf{AcceptedRequest} \subs \lf{Request}\\
      \lf{RefusedRequest} \subs \lf{Request} \sqcap \non \lf{AcceptedRequest}\\[1ex]
      
      \lf{ProduceRequest} \subs \lf{Request}\\
      \lf{AcceptedProduceRequest} \subs \lf{ProduceRequest} \sqcap \lf{AcceptedRequest}\\[1ex]
      
      \lf{ShippingRequest} \subs \lf{Request}\\
      \lf{AcceptedShippingRequest} \subs \lf{ShippingRequest} \sqcap \lf{AcceptedRequest}\\[1ex]
      
      \lf{ProduceOffer} \subs \lf{Offer}\\
      \lf{ShippingOffer} \subs \lf{Offer}
  \end{array}
  $
  \end{center}
  The theory states that a request can be classified as accepted or refused by
  one of the two agents: we further characterize offers, requests and accepted
  requests by the agent to which they refer.  To relate requests to offers and
  to the information that they convey, we include in $\Tbox$ 
  the following axioms:
  \begin{center}\small
  $
  \begin{array}{l}
      \lf{Offer} \subs \forall \lf{hasCost}.\lf{Price}\\
      \lf{Request} \subs \forall \lf{hasOffer}.\lf{Offer}\\[1ex]
      
      \lf{ShippingRequest} \subs \forall \lf{hasDestination}.\lf{Location}\\
      \lf{ProduceRequest} \subs \forall \lf{hasProduct}.\lf{Product}
  \end{array}
  $
  \end{center}  
  In other words, every offer in $\lf{Offer}$ specifies its $\lf{Price}$ by the role
  $\lf{hasCost}$; requests relate to their offers by the role $\lf{hasOffer}$;
  finally, a $\lf{ShippingRequest}$ contains information about the
  $\lf{Location}$ to where to ship by the role $\lf{hasDestination}$ and a
  $\lf{ProduceRequest}$ describes the $\lf{Product}$ to buy by the role
  $\lf{hasProduct}$.

  Given a finite set of individual names $\Ncal$, we assume to have a suitable
  $\eta \in \ity_\Ncal(\Tbox)$ justifying the validity of $\Tbox$ with respect
  to elements of $\Ncal$.  Note that $\Tbox$ only represents a TBox, thus
  information terms of its subsumptions are functions mapping information terms
  of the included concept in those of the including concept. If we assume to
  store assertions of an ABox over $\Ncal$ in some kind of database (e.g., a
  relational database or the data part of a logic program), the functions for
  each of these information terms can be implemented as query prototypes (to be
  instantiated with individuals of $\Ncal$) over the database.
\EndEs
\end{example}

\noindent
Given a finite subset $\Ncal$ of $\NI$, an {\em $\Ncal$-substitution} $\s$ is a
map $\s : \VAR \to \Ncal$.  We extend $\s$ to $\Lcal_\Ncal$ as usual: if
$c\in\Ncal$, $\s c = c$; for a formula $K$ of $\Lcal_\Ncal$, $\s K$ denotes the
closed formula of $\Lcal_\Ncal$ obtained by replacing every variable $x$
occurring in $K$ with $\s(x)$; given a set of formulas $\G$, $\s\G$ is the set
of $\s K$ such that $K\in \G$. If $c\in\Ncal$, $\s[c/p]$ is the
$\Ncal$-substitution $\s'$ such that $\s'(p)=c$ and $\s'(x)=\s(x)$ for $x\neq
p$. A $\Ncal$-substitution $\s$ is a \emph{closing substitution} for a set of
formulas $\G$ if $\s\G$ is a set of closed formulas.

Now, let us consider the natural deduction calculus $\CALC$ for $\BCDLZ$ whose
rules are given in Figure~\ref{table:rules1}. We denote with $\pi::\G\seq K$ the
fact that $\pi$ is a proof of $\G\seq K$ and with $\G\dimBCDLZ K$ the fact that
there exists a proof $\pi::\G\seq K$ in $\CALC$.  For a detailed presentation of
the calculus and its properties we refer the reader to
\cite{FerFioFio:2010}. 
Here we only note that $\CALC$ is sound with respect to
the information term semantics, namely:

\begin{theorem}[Soundness]
  Let $\G\cup\{K\}\subseteq \Lcal_\Ncal$, let $\pi::\G\seq K$ be a proof of
  $\CALC$ and let $\Sigma$ be the set of all the closing $\Ncal$-substitutions
  for $\G\cup\{K\}$. Then there exists an operator
  \[
    \Phi^\pi_\Ncal:\bigcup_{\s\in\Sigma}\ity_\Ncal(\s\G)\to\bigcup_{\s\in\Sigma}\ity_\Ncal(\s K)
  \]
  such that, for every $\ov{\g}\in\ity(\s\Gamma)$ and for every model $\Mcal$
  for $\Lcal_\Ncal$, $\Mcal\real{\ov{\g}}\s\Gamma$ implies
  $\Mcal\real{\extract{\pi}{\ov{\g}}}\s K$. \qed
\end{theorem}

\noindent
We remark that the proof of the above theorem is constructive. As shown
in~\cite{FerFioFio:2010} we can effectively extract from the proof $\pi$ the
operator $\Phi^\pi_\Ncal$. This plays an important role in the definition of our
service composition calculus in Section~\ref{sec:papdl}.

\begin{figure}[ht]
  \begin{center}
    {\small
      \[
      \begin{array}{c}
        \hline\\
        \begin{prooftree}
          \G_1\seq t:C \hspace{2ex}\G_2\seq t:\non C
          \mbox{} \justifies \G_1,\G_2\seq \bot \nome{\bot I}
        \end{prooftree}
        \hspace{3ex}
        \begin{prooftree}
          \G\seq \bot \justifies \G\seq K \nome{\bot E}
        \end{prooftree}
        \hspace{4ex}
        \begin{prooftree}
          \G\seq t:A
          \mbox{} \justifies \G,A\im C\seq t:C \nome{\im E}
        \end{prooftree}
        \hspace{4ex}
        \begin{prooftree}
          \G, t:C \seq \bot \justifies \G\seq t:\non C \nome{\non I}
        \end{prooftree}
        \\[4ex]
        \begin{prooftree}
          \G_1\seq t:C_1\hspace{2ex}\G_2\seq t:C_2
          \justifies \G_1,\G_2 \seq t: C_1\sqcap C_2 \nome{\sqcap I}
        \end{prooftree}
        \hspace{8ex}
        \begin{prooftree}
          \G\seq t:C_1\sqcap C_2 \justifies \G\seq t:C_k \nome{\sqcap
            E_k}~k\in \{1,2\}
        \end{prooftree}
        \\[4ex]
        \begin{prooftree}
          \G\seq t:C_k \justifies \G\seq t:C_1\sqcup C_2 \nome{\sqcup
            I_k}~k\in \{1,2\}
        \end{prooftree}
        \hspace{6ex}
        \begin{prooftree}
          \G_1\seq t:C_1\sqcup C_2\hspace{2ex}
          \G_2, t:C_1 \seq K\hspace{2ex}
          \G_3\seq t:C_2 \seq K \justifies \G_1,\G_2,\G_3\seq K \nome{\sqcup E}
        \end{prooftree}
        \\[4ex]
        \begin{prooftree}
          \G\seq u:C \justifies \G, (t,u):R \seq t: \exists R.C
          \nome{\exists I}
        \end{prooftree}
        \hspace{4ex}
        \begin{prooftree}
          \G_1\seq t: \exists R.C\hspace{2ex} \G_2,(t,p):R,p:C\seq K
          \justifies \G_1,\G_2\seq K \nome{\exists E}
        \end{prooftree}\hspace{2ex}
        \mbox{\begin{minipage}{16ex}
            \scriptsize where $p$ does not occur in $\G_2\cup\{K\}$ and 
            $p\neq t$
          \end{minipage}}
        \\[4ex]
        \begin{prooftree}
          \G,(t,p):R \seq p:C \justifies \G\seq t:\forall R.C
          \nome{\forall I}
        \end{prooftree}\hspace{2ex}
        \mbox{\begin{minipage}{22ex}
            \scriptsize where $p$  does not occur in $\G$
            and $p\neq t$
          \end{minipage}} \hspace{8ex}
        \begin{prooftree}
          \G \seq s:\forall R.C \justifies \G, (s,t):R  \seq t:C
          \nome{\forall E}
        \end{prooftree}
        \\[4ex]
        \hline
      \end{array}
      \]
    }
  \end{center}
  \caption{The rules of the calculus $\CALC$}
   \label{table:rules1}
\end{figure}


\section{Service Specifications}
\label{sec:serviceSpecifications}

In this section we introduce the basic definitions for the description of
systems and for the specification of services operating on them.  A
\emph{service specification (over $\Lcal_\Ncal$)} is an expression of the form
$\SERV{s}{x}{P}{Q}$ where: $\serv{s}$ is a label that identifies the service;
$x$ is the input parameter of the service (to be instantiated with an individual
name from $\Ncal$); $P$ and $Q$ are concepts over $\Lcal_\Ncal$. $P$ is called
the service \emph{pre-condition}, denoted with $\pre(\serv{s})$, and $Q$ the
service \emph{post-condition}, denoted with $\post(\serv{s})$.  Given a service
specification $\SERV{s}{x}{P}{Q}$ over $\Lcal_\Ncal$ we call 
\emph{service implementation} a function
\[
\Phi_s:\bigcup_{t\in\Ncal}\ity_\Ncal(t:P)\to \bigcup_{t\in\Ncal}\ity_\Ncal(t:Q)
\]
We denote with the pair $(\SERV{s}{x}{P}{Q},\Phi_s)$ (or simply with
$(\serv{s},\Phi_s)$) a \emph{service definition} over $\Lcal_\Ncal$.

Essentially, a service definition corresponds to an effective Web service. The
service specification provides the formal description of the behavior of the
service in terms of pre- and post- conditions. The function $\Phi_s$ represents a
formal description of service implementation (i.e., of the input/output
function).

The notion of correctness is modeled as follows. Given a language $\Lcal_\Ncal$,
a service definition $(\SERV{s}{x}{P}{Q},\Phi_s)$ over $\Lcal_\Ncal$ and a
model $\Mcal$ for $\Lcal_\Ncal$, $\Phi_s$ \emph{uniformly solves}
$\SERV{s}{x}{P}{Q}$ in $\Mcal$ iff, for every individual name $t\in\Ncal$ and
every $\a \in \ity_\Ncal(t:P)$ such that $\Mcal \real{\a} t:P$, $\Mcal
\real{\Phi_s(\a)} t:Q$.

\begin{example}[Service specification]
  We can now model the services provided by the Producer and Shipper agents.
  \begin{quote}\small
    \begin{tabbing}
      \hspace{3ex}\=\hspace{2ex}\=\hspace{2ex}\=\hspace{2ex}\=\kill
      $\serv{DoProduceRequest}(\lf{req})::$\\
      \>$\lf{ProduceRequest} \,\e\, \exists \lf{hasProduct.Product}$\\
      \>$\PAPrule \lf{RefusedRequest} \,\o\, (\, \lf{AcceptedProduceRequest} \,\e\,$\\
      \>\>$\exists\lf{hasOffer.}(\, \lf{ProduceOffer} \,\e\, \exists \lf{hasCost.Price} 
      \,)\,)$\\[3ex]
      
      $\serv{DoShippingRequest}(\lf{req})::$\\
      \>$\lf{ShippingRequest} \,\e\, \exists \lf{hasDestination.Location}$\\
      \>$\PAPrule \lf{RefusedRequest} \,\o\, (\, \lf{AcceptedShippingRequest} \,\e\,$\\
      \>\>$\exists\lf{hasOffer.}(\, \lf{ShippingOffer} \,\e\, \exists \lf{hasCost.Price} 
      \,)\,)$
    \end{tabbing}
  \end{quote}
  The service described by $\serv{DoProduceRequest}$ takes as input a request
  $\lf{req}$ specifying the required product and must classify it according to
  the service post-condition: namely, the service can answer with a refusal to
  the request (by classifying $\lf{req}$ in $\lf{RefusedRequest}$) or it can
  accept the request and produce an offer with a price specified by the
  $\lf{hasCost}$ role. The $\serv{DoShippingRequest}$ service works in a
  similar way: it takes as input the destination where to ship the product and
  either refuses the request or it accepts the request providing an offer with
  the associated price.  
  
  In our setting, service implementations correspond to functions mapping
  information terms for the pre-condition into information terms for the
  post-condition. These functions formalize the behavior of the effective
  implementation of the web services.  In particular let us consider the
  implementation $\Impl{DPR}$ of the $\serv{DoProduceRequest}$ service. Let
  $\lf{req\_1}$ be the individual name representing a request. The input of
  $\Impl{DPR}$ is any information term for $\a\in
  \ity_\Ncal(\lf{req\_1}:\pre(\serv{DoProduceRequest}))$.  \lf{req\_1} can be
  seen as a reference to a database record providing the information required by
  the service precondition and $\a$ can be seen as a structured representation
  of such information.
  Let us suppose that $\a = (\true,
  (\lf{book\_1}, \true))$; this information term means that \lf{req\_1} is a
  product request with associated product $\lf{book\_1}$. Now, let
  $\b=\Impl{DPR}(\a)\in\ity(\lf{req\_1}:\post(\serv{DoProduceRequest}))$. If
  $\b=(1, \true)$, this classify $\lf{req\_1}$ as refused. Otherwise $\b$ could
  be $(2, (\true, (\lf{off\_1}, (\true, (\lf{price\_1}, \true) ) ) ) )$ which
  classifies $\lf{req\_1}$ as accepted and specifies 
  that there is an offer $\lf{off\_1}$ with associated
  price $\lf{price\_1}$ for the requested product. The implementation
  $\Impl{DSR}$ of $\serv{DoShippingRequest}$ acts in a similar way.
    
  To conclude, we remark that the intended model $\Mcal$ we use to evaluate the
  correctness of the system is implicitly defined by the knowledge base of the
  system. Indeed, $\Mcal\real{\a}{\lf{req\_1}:\pre(\serv{DoProduceRequest})}$ if
  and only if in our system $\lf{req\_1}$ effectively codify a request and
  \lf{book\_1} is classified as a product. In this case, since $\Impl{DPR}$
  uniformly solves the service specification, we know that
  $\Mcal\real{\b}{\lf{req\_1}:\post(\serv{DoProduceRequest})}$: this trivially
  corresponds to the fact that, looking at its knowledge base, the Producer can
  generate its offer.
  \EndEs
\end{example}

\noindent
The problem of service composition amounts to build a new service from a family
of implemented services. We formalize this problem in the context of an
\emph{environment}, that is a structure $ \Env = \stru{\Lcal_\Ncal, \Tbo, \eta,
  (s_1,\Phi_1),\dots,(s_n,\Phi_n)} $ where:
\begin{itemize}[leftmargin=*]
\item $\Tbo$ is a theory over the language $\Lcal_\Ncal$;
\item $\eta\in\ity_\Ncal(\Tbo)$;
\item for every $i\in\{1,\dots,n\}$, $(s_i,\Phi_i)$ is a service definition in
  $\Lcal_\Ncal$.
\end{itemize}
Given a model $\Mcal$ for $\Lcal_\Ncal$ we say that $\Mcal$ is a model for
$\Env$ iff $\Mcal\real{\eta}\Tbo$ and for every $i\in\{1,\dots, n\}$, $\Phi_i$
uniformly solves $s_i$ in $\Mcal$.

A service specification $s'$ is \emph{solvable} in $\Env$ if there exists an
implementation $\Phi'$ of $s'$ such that, for every model $\Mcal$ of
$\Lcal_\Ncal$, if $\Mcal$ is a model for $\Env$ then $\Phi'$ uniformly
solves $s'$ in $\Mcal$.

\begin{example}[Composition problem definition]\label{ex:environment}
  Given the previous specifications, we are now ready to state the composition
  problem.  We want to combine the services $\serv{DoProduceRequest}$ and
  $\serv{DoShippingRequest}$ to provide the User with a single service to
  request both the production and the delivery of an object. To do this, we
  define a third service that composes the offers from the two agents:
  \begin{center}\small
    \begin{tabbing}
      \hspace{3ex}\=\hspace{2ex}\=\hspace{2ex}\=\hspace{2ex}\=\kill
      $\serv{ProcessOffers}(\lf{req})::$\\
      \>$\lf{AcceptedProduceRequest} \,\e\,
      \exists\lf{hasOffer.}(\, \lf{ProduceOffer} \,\e\, 
      \exists \lf{hasCost.Price} \,) \;\e$\\
      \>$\lf{AcceptedShippingRequest} \,\e\,
      \exists\lf{hasOffer.}(\, \lf{ShippingOffer} \,\e\, \exists \lf{hasCost.Price} \,)$\\
      \>$\PAPrule \lf{AcceptedRequest} \,\e\,
      \exists\lf{hasOffer.}(\, \lf{Offer} \,\e\, \exists \lf{hasCost.Price} 
      \,)$
    \end{tabbing}
  \end{center}
  Let $\Impl{PO}$ be the implementation of $\serv{ProcessOffers}$. We define the
  environment $\EnvPS = \stru{\Ln, \Tbox, \eta, S_1, S_2, S_3}$ where
  $S_1=(\serv{DoProduceRequest},\Impl{DPR})$,
  $S_2=(\serv{DoShippingRequest},\Impl{DSR})$ and $S_3=(\serv{ProcessOffers},
  \Impl{PO})$.  The problem can be now reduced to the definition of a suitable
  service specification that is solvable in such environment. \EndEs
\end{example}

\noindent
Now, the main point of service composition is to effectively build the
implementation of the service specification starting from the environment.  This
problem can be solved in two ways: the first solution consists in the
definition of a composition language which allows the user to build up a new
service starting from the environment. The second is given by providing a method to
automatically build up the new service implementation.

The formalization of the composition problem in the framework of a
(constructive) logic allows to use the proof-theoretical properties of the
logical system to support the composition problem.  In this paper we concentrate
on the definition of a composition language.  As for the problem of automatic
service composition, it can be seen as a reformulation of the
\emph{program-synthesis} problem, a problem which has a long tradition in the
constructive logics context and which has already been studied in the framework of
$\BCDL$, see~\cite{BozFerFioFio:07,FerFioFio:2010}.


\section{Composition Calculus \PAPdl}
\label{sec:papdl}

The composition calculus we describe in this section is inspired by
PAP~\cite{MigMosOrn:91}, a calculus which support program synthesis from
proofs of a constructive logical system. Our calculus allows to manually compose
services guaranteeing the correctness of the composed service. The main advantage
of our formalization is that service composition can be supported by an
appropriate proof-system.  This tool can be used to check the correctness of
rule applications and to automatically build the proofs of the applicability
conditions. 

A \emph{composition} over an environment $\Env = \stru{\Lcal_\Ncal, \Tbo, \eta,
  (s_1,\Phi_1),\dots,(s_n,\Phi_n)}$ is defined as:\\[-4ex]
\begin{center}\small
  $\Tab{\SERV{s}{x}{P}{Q}}
  {
    \begin{array}{c}
      \Pi_1 : \SERV{s_1}{x}{P_1}{Q_1}\\
      \cdots\\
      \Pi_n : \SERV{s_n}{x}{P_n}{Q_n}
    \end{array}
  }{\footnotesize $r$}$
\end{center}
where: 
\begin{itemize}[leftmargin=*]
\item $\SERV{s}{x}{P}{Q}$ is a service specification over $\Env$;
\item $r$ is one of the rules of the composition calculus $\PAPdl$;
\item For every $i\in\{1,\dots,n\}$, $\Pi_i:\SERV{s_i}{x}{P_i}{Q_i}$ is a
  service composition over $\Env$ that meets the applicability conditions of
  $r$.
\end{itemize}

\noindent
The rules of the composition calculus \PAPdl and their computational
interpretation $\Phi_s$ are given in Figure~\ref{fig:pap-dl}.  In the rules, the
service specification $\SERV{s}{x}{P}{Q}$ is called the \emph{main sequent} of
the rule and represents the specification of the service to be composed.  The
service specifications $\SERV{s_i}{x}{P_i}{Q_i}$ are called \emph{subsequents}
of the rule and represent the services involved in the composition.  The
sequents must satisfy the \emph{applicability conditions (AC)} of the
rule. These conditions describe the role of the subsequents in the composition
of the main sequent: in order to verify the correctness of compositions, the
proof checker must verify the truth of such conditions.

\begin{figure}[t]\small
  \[
  \begin{array}{l@{\hspace{4ex}}l}
    \hline\\
    \fbox{\begin{minipage}{25ex}
        $\Tab{\SERV{s}{x}{A}{B}}
        {
          \begin{array}{c}
            \;\; \SERV{s_1}{x}{A_1}{B_1}\\
            \cdots\\
            \;\; \SERV{s_n}{x}{A_n}{B_n}
          \end{array}}{AND}$
      \end{minipage}}&
    \begin{minipage}{50ex}
      $\mbox{AC}~\left\{
        \begin{array}{cl}
          (a_k) & \Tbo,x:A \dimBCDLZ x:A_k,\, \mbox{for $k\in\{1,\dots,n\}$}\\[1ex]
          (b)   & \Tbo, x:B_1 \e \ldots \e B_n \dimBCDLZ x:B
        \end{array}\right.$\\[2ex]
      $\Phi_{s}(\a)=\Phi_{b}(\Phi_{s_1}(\Phi_{a_1}(\a)), \ldots,\Phi_{s_n}(\Phi_{a_n}(\a)))$
    \end{minipage}\\[9ex]
    \fbox{\begin{minipage}{25ex}
        $\Tab{\SERV{s}{x}{A}{B}}
        {
          \begin{array}{c}
            \;\; \SERV{s_1}{x}{A_1}{B_1}\\
            \cdots\\
            \;\; \SERV{s_n}{x}{A_n}{B_n}
          \end{array}}{CASE}$
      \end{minipage}}&
    \begin{minipage}{50ex}
      $\mbox{AC}~\left\{
        \begin{array}{cl}    	
          (a) & \Tbo, x:A \dimBCDLZ x:A_1 \o \ldots \o A_n\\[1ex]
          (b_k) & \Tbo, x:B_k \dimBCDLZ x:B,\, \mbox{for $k\in\{1,\dots,n\}$}
        \end{array}\right.$\\[2ex]
      $\Phi_{s}(\a)\;=\; \Phi_{b_k}(\,\Phi_{s_k}(\a_k) \,)$ where $(k, \a_k) = \Phi_a(\a)$
    \end{minipage}\\[8ex]
    \fbox{\begin{minipage}{25ex}
        $\Tab{\SERV{s}{x}{A}{B}}
        {\begin{array}{c}
            \;\; \SERV{s_1}{x}{A_1}{B_1}\\
            \cdots\\
            \;\; \SERV{s_n}{x}{A_n}{B_n}
          \end{array}}{SEQ}$
      \end{minipage}} &
    \begin{minipage}{50ex}
      $\mbox{AC}~\left\{
        \begin{array}{cl}
          (b_1) & \Tbo, x:A \dimBCDLZ x:A_1\\[1ex]
          (b_k) & \Tbo, x:B_{k-1} \dimBCDLZ x:A_k,\, \mbox{for $k\in\{2,\dots,n\}$}\\[1ex]
          (c) & \Tbo, x:B_n \dimBCDLZ x:B
        \end{array}\right.$\\[2ex]
      $\Phi_{s}(\a) \;=\; \Phi_{c}(\, \Phi_{s_n} \cdot \Phi_{b_n} \cdot
      \,\ldots\, \cdot \Phi_{s_1} \cdot \Phi_{b_1} (\a) \,)$
    \end{minipage}\\[10ex]
    \fbox{
      \begin{minipage}{23ex}
        $\SERV{s}{x}{A}{B}\quad\mbox{\small AX}$
      \end{minipage}
    } &
    \begin{minipage}{50ex}
      $\mbox{AC}\hspace{1em}(a)~\Tbo,x:A \dimBCDLZ x:B$\\[1ex]
      $\Phi_{s}(\a) \;=\; \Phi_{a}(\a)$
    \end{minipage}\\[6ex]
    \fbox{
      \begin{minipage}{23ex}
        $\SERV{s}{x}{A}{B}\quad\mbox{\small ENV}$
      \end{minipage}
    } & 
    \mbox{with $(\serv{s},\Phi_s)$ a service defined in $\Env$}\\[3ex]
    \hline
  \end{array}
  \]
  \caption{The rules of calculus \PAPdl}
  \label{fig:pap-dl}
\end{figure}
\noindent
The composition rules have both a logical and a computational reading.  Given a
service composition $\Pi$ with main sequent $\SERV{s}{x}{P}{Q}$, we define the
function $\Phi_{\serv{s}}: \bigcup_{t\in\Ncal}\ity_\Ncal(t:P) \to
\bigcup_{t\in\Ncal}\ity_\Ncal(t:Q)$ associated with $\serv{s}$. The function is
inductively defined on the last rule $r$ applied in $\Pi$. Here we assume the
following conventions: given a subsequent $s'$ of the rule $r$, we denote with
$\Phi_{s'}$ its computed function; given the applicability condition
$(a)~\G\dimBCDLZ x:A$ of the rule $r$ we denote with $\Phi_{a}$ the operator
corresponding to the proof $\pi::\G\seq x:A$ defined according to
Section~\ref{sec:logic}.

Inspecting the rules of Figure~\ref{fig:pap-dl} we see that: 
\begin{itemize}[leftmargin=*]
\item 
  The AND rule represents a $\sqcap$ introduction on the right hand side
  of the specification sequents: the services composed by this rule are seen as
  a parallel execution of the sub services.
\item 
  The CASE rule represents a $\sqcup$ elimination on the left hand side of
  the specification sequent: the services composed by this rule are seen as in a
  case construct, in which the applicability condition determines the executed
  sub-service.
\item 
  The SEQ rule represents a composition given as a sequential execution of
  the sub-services and a composition of proofs under the logical reading.
\item 
  The AX rule states that the system can infer specifications provable
  under a suitable calculus for $\BCDLZ$.
\item 
  The ENV rule allows to use the specifications given in the environment $\Env$.
\end{itemize}

\noindent
Let us complete our example with a sample service composition.

\begin{example}[Service Composition]
  Given the environment $\EnvPS$ defined in Example~\ref{ex:environment} and the
  rules of $\PAPdl$, we can define a new service $\serv{ProduceAndShip}$ as the 
  composition $\Pi$ of the stated specifications as follows:
  \begin{center}\footnotesize
    \begin{tabbing}
      \hspace{6ex}\=\hspace{6ex}\=\hspace{6ex}\=\hspace{2ex}\=\hspace{2ex}\=\hspace{2ex}\=\kill
      $\serv{ProduceAndShip}(\lf{req})::$\\
      \>$\lf{ProduceRequest} \e \lf{ShippingRequest} \e$\\
      \>$\exists \lf{hasProduct.Product} \e \exists \lf{hasDestination.Location} \PAPrule$\\[.5ex]
      \>$\lf{RefusedRequest} \,\sqcup\, (\,\lf{AcceptedRequest} \,\e\,
      \exists \lf{hasOffer.}(\,\lf{Offer} \,\e\, \exists \lf{hasCost.Price}\,)\,)$\\
      \>\RuleLine{72ex}{SEQ}\\
      \>$\Pi_1:\serv{DoRequest}(\lf{req})::$\\
      \>\>$\lf{ProduceRequest} \e \lf{ShippingRequest} \e$\\
      \>\>$\exists \lf{hasProduct.Product} \e \exists 
      \lf{hasDestination.Location} \PAPrule$\\[.5ex]
      \>\>$\lf{RefusedRequest} \,\sqcup\,(\,(\,\lf{AcceptedProduceRequest} \,\e\,$\\
      \>\>$\exists \lf{hasOffer.}(\,\lf{ProduceOffer} \,\e\, \exists \lf{hasCost.Price}\,)\,)$\\
      \>\>$\,\e\, (\,\lf{AcceptedShippingRequest} \,\e\,$\\
      \>\>$\exists \lf{hasOffer.}(\,\lf{ShippingOffer}\,\e\,\exists 
      \lf{hasCost.Price}\,)\,)\,)$\\
      \>\>\RuleLine{64ex}{AND}\\
      \>\>$\serv{DoProduceRequest}(\lf{req})::$\hspace{28ex}{ENV}\\
      \>\>\>$\lf{ProduceRequest} \,\e\, \exists \lf{hasProduct.Product} \PAPrule$\\[.5ex]
      \>\>\>$\lf{RefusedRequest} \,\o\, (\, \lf{AcceptedProduceRequest} \,\e\,$\\
      \>\>\>$\exists\lf{hasOffer.}(\, \lf{ProduceOffer} \,\e\, 
      \exists \lf{hasCost.Price}\,)\,)$\\[1ex]
      \>\>$\serv{DoShippingRequest}(\lf{req})::$\hspace{28ex}{ENV}\\
      \>\>\>$\lf{ShippingRequest} \,\e\, \exists \lf{hasDestination.Location} \PAPrule$\\[.5ex]
      \>\>\>$\lf{RefusedRequest} \,\o\, (\, \lf{AcceptedShippingRequest} \,\e\,$\\
      \>\>\>$\exists\lf{hasOffer.}(\, \lf{ShippingOffer} \,\e\, 
      \exists \lf{hasCost.Price}\,)\,)$\\[2ex]
      \>$\Pi_2:\serv{PresentOffer}(\lf{req})::$\\
      \>\>$\lf{RefusedRequest} \,\sqcup\,(\,(\,\lf{AcceptedProduceRequest} \,\e\,$\\
      \>\>$\exists \lf{hasOffer.}(\,\lf{ProduceOffer} \,\e\, \exists \lf{hasCost.Price}\,)\,)$\\
      \>\>$\,\e\, (\,\lf{AcceptedShippingRequest} \,\e\,$\\
      \>\>$\exists \lf{hasOffer.}(\,\lf{ShippingOffer} \,\e\, 
      \exists \lf{hasCost.Price}\,)\,)\,) \PAPrule$\\[.5ex]
      \>\>$\lf{RefusedRequest} \,\sqcup\, (\,\lf{AcceptedRequest} \,\e\,$\\
      \>\>$\exists \lf{hasOffer.}(\,\lf{Offer} \,\e\, \exists \lf{hasCost.Price}\,)\,)$\\
      \>\>\RuleLine{64ex}{CASE}\\
      \>\>$\serv{RefuseRequest}(\lf{req})::$\hspace{30ex}{AX}\\
      \>\>\>$\lf{RefusedRequest}	\PAPrule \lf{RefusedRequest}$\\[1ex]
      \>\>$\serv{ProcessOffers}(\lf{req})::$\hspace{32ex}{ENV}\\
      \>\>\>$\lf{AcceptedProduceRequest} \,\e\,$\\
      \>\>\>$\exists\lf{hasOffer.}(\, \lf{ProduceOffer} \,\e\,
      \exists \lf{hasCost.Price} \,) \;\e$\\
      \>\>\>$\lf{AcceptedShippingRequest} \,\e\,$\\
      \>\>\>$\exists\lf{hasOffer.}(\, \lf{ShippingOffer} \,\e\,
      \exists \lf{hasCost.Price} \,)\PAPrule$\\[.5ex]
      \>\>\>$\lf{AcceptedRequest} \,\e\,
      \exists\lf{hasOffer.}(\, \lf{Offer} \,\e\, \exists \lf{hasCost.Price}\,)$
    \end{tabbing}
  \end{center}
  The behavior of this service is defined as follows: using the
  $\serv{DoRequest}$ service (service composition $\Pi_1$), it first invokes the
  $\serv{DoProduceRequest}$ and the $\serv{DoShippingRequest}$ services to query
  the Producer and the Shipper over the combined request $\lf{req}$.  The answer
  of the two is then combined by $\serv{PresentOffer}$ (service composition
  $\Pi_2$): by a case construct, this sub-service either responds that the
  request $\lf{req}$ has been classified as refused, or it accepts the request
  and generate the combined price using the $\serv{ProcessOffers}$ service.
  
  Let us discuss how the composite service computes information terms by
  explaining a sample execution. Let $\lf{req\_2}$ be both a $\lf{ProduceRequest}$ and
  a $\lf{ShippingRequest}$ with associated product $\lf{book\_1}$ and shipping
  destination $\lf{my\_home}$. Then, a call of $\serv{ProduceAndShip}$ over
  $\lf{req\_2}$ has as input information term $\a_ 1 = ( \true, ( \true, (
  (\lf{book\_1}, \true), (\lf{my\_home}, \true) ) ) )$.
  Following the composition, 
  The execution of $\serv{ProduceAndShip}$ starts with the sequence 
  construct and the first invoked service is $\serv{DoRequest}$
  which process the information term $\a_1$.
  $\serv{DoRequest}$ consists of a parallel call to the services of the Producer
  and the Shipper.  The request to the Producer is executed as a call to
  $\serv{DoProduceRequest}$. According to the conditions of the AND rule,
  we have a proof:
  \[
    \pi_1:: \Tbox, x: \pre(\serv{DoRequest}) \dimBCDLZ x:\pre(\serv{DoProduceRequest})
  \]
  The corresponding operator $\Phi^{\pi_1}_\Ncal$ allows us to extract from
  $\a_1$ the information term $( \true, (\lf{book\_1}, \true)
  )\in\ity_\Ncal(\lf{req\_2}:\pre(\serv{DoProduceRequest}))$. Let us suppose
  that the Producer accepts the request and produces an offer $\lf{p\_off}$ with
  an associated price. The offer is codified in the information term
  $\a_2=\Phi_{DPR}(( \true, (\lf{book\_1}, \true) )$. Let us assume that
  $\a_2$ has the following form:
  \[
  \a_2= (2, ( \true, ( \lf{p\_off}, ( \true, ( \lf{p\_off\_price}, \true ) ) ) ) )
  \]
  The request to the Shipper consists in a call to $\serv{DoShippingRequest}$
  with input information term $( \true, (\lf{my\_home}, \true) )$. 
  Also in this
  case this information term is generated from the operator associated with an
  applicability rule.  
  As above, if the Shipper accepts the
  request with an offer $\lf{s\_off}$ and its price, then the output 
  information term is:
  \[
  \a_3=(2, ( \true, ( \lf{s\_off}, ( \true, ( \lf{s\_off\_price}, \true ) ) ) ) )
  \]
  Now the applicability conditions of the AND composition rule, 
  in particular the proof:
  \begin{eqnarray*}
    \pi_2:: &&\Tbox, x:\post(\serv{DoProduceRequest}) \e \post(\serv{DoShippingRequest})\\
    &&\dimBCDLZ
    x: \post(\serv{DoRequest})
  \end{eqnarray*}
  allows us to combine $\a_2$ and $\a_3$ to get
  an $\a_4 \in \ity_\Ncal(\lf{req\_2}:\post(\serv{DoRequest}))$ as follows:
  \begin{center}\small
    $
    \a_4 = (2, (( \true, ( \lf{p\_off}, ( \true, ( \lf{p\_off\_price}, \true ) ) ) ),
    ( \true, ( \lf{s\_off}, ( \true, ( \lf{s\_off\_price}, \true ) ) ) ) 
    ) )
    $
  \end{center}
  Proceeding in the sequence, the previous responses are combined by a call of
  $\serv{PresentOffer}$ with input information term $\a_4$.
  By the AC of the CASE
  construct, as the request has been accepted by both agents, we enter in the
  second of the cases and we call $\serv{ProcessOffers}$ with input information
  term: 
  \begin{center}\small
    $\a_5=
    (\, ( \true, ( \lf{p\_off}, ( \true, ( \lf{p\_off\_price}, \true ) ) ) ),
    ( \true, ( \lf{s\_off}, ( \true, ( \lf{s\_off\_price}, \true ) ) ) ) \,)
    $
  \end{center}
  The service combines the offers producing a composite offer $\lf{ps\_off}$
  with its associated price \lf{ps\_off\_price} modeled by the information term: 
  \begin{center}\small
    $
    ( \true, ( \lf{ps\_off}, ( \true, ( \lf{ps\_off\_price}, \true ) ) ) )
    $
  \end{center}
  Finally the output  of $\serv{PresentOffer}$ and $\serv{ProduceAndShip}$ is:
  \[
  (2, ( \true, ( \lf{ps\_off}, ( \true, ( \lf{ps\_off\_price}, \true ) ) ) ) )
  \]
  This object states that the request has been accepted and it contains both the
  object representing the composite offer ($\lf{ps\_off}$) and its composite
  price ($\lf{ps\_off\_price}$).\EndEs
\end{example}

\noindent
To conclude this section we state the result asserting the soundness of the
rules with respect to uniform solvability. Its proof easily follows by induction
on the structure of the composition $\Pi$.

\begin{theorem}
  Let $\Env = \stru{\Lcal_\Ncal, \Tbo, \eta, (s_1,\Phi_1),\dots,(s_n,\Phi_n)}$
  be an environment and let $\SERV{s}{x}{P}{Q}$ be the main sequent of a
  composition $\Pi$ over $\Env$. For every model $\Mcal$ for $\Lcal$, if $\Mcal$
  is a model for $\Env$ then the function $\Phi_s$ extracted from $\Pi$
  uniformly solves  $s$ in $\Mcal$.\qed
\end{theorem}


\section{Related Works and Conclusions}
\label{sec:conclusion}

In this section we review some of the current approaches for the composition of
semantic Web services and we discuss the relations with our proposal.

One of the most relevant proposals for the semantic description
of Web services is OWL-S~\cite{owls:2007}.
An OWL-S service is described by three representations conforming to three
distinct ontologies: 
a \emph{service profile}, describing its use in terms of inputs, outputs, pre-
and post- conditions, a \emph{process model}, stating the flow of interactions
that composes the service, and a \emph{grounding model}, describing the details
about its interface.  Among these, the profile and the process model define an
abstract representation of the service. In particular, the service profile
gives a declarative description of the service. The process model describes the
possible interactions with a service, that is its composition with other given
services.  The model mainly distinguish services into \emph{atomic} and
\emph{composite processes}: atomic processes do not have a representation of
their internal structure and thus they are entirely defined by their profiles;
on the other hand, composite processes are representations of compositions of
processes linked by some control structure.

Many tasks can be supported by a pure atomic view of the services: one of these
tasks is planning, which can be seen as a way to synthesize service
compositions. In~\cite{owls:2007} it is stated that the execution behavior of
control constructs of the composite processes can not be suitably represented in
OWL-DL: this brought to several translations of OWL-S compositions into
different formalisms.
The idea is that by a formalization of control structures
one can reason over the definition of composite processes.
One of these formalizations has been presented in~\cite{NarayananM:02}: the
authors define a translation from DAML-S descriptions to Petri Nets and give
procedures to compute composition, verification and simulation of services.
However, as noted in~\cite{TraversoP:04}, this approach only allows sequential
composition of atomic services.
A similar approach is presented in~\cite{TraversoP:04} in which OWL-S process
models are encoded into state transition systems: synthesis of compositions is
performed by planning techniques.  This approach features the representation of
non deterministic outcomes of services, complex specifications of goals and the
translation of the resulting plans to executable processes. We must note that
the above approaches are mostly based on the process representation of services.

An approach that highlights the relationships between composition and software
synthesis is proposed in~\cite{MatskinRao:02}.  This approach composes services
on the base of their service profiles.
The idea is that Web services can be treated as software components, thus
service composition can be carried out as a problem of software composition.
The approach uses the Structural Synthesis Program (SSP)
method~\cite{MatskinT:01} to extract compositions.
We notice that SSP is based on the implicative part of the intuitionistic
propositional calculus and it can only define sequential or conditional
compositions in general.

A different representation of semantic Web services is given by the WSMO
ontology~\cite{BruijnLPF:06} and its Web service modeling language WSML. As
OWL-S, these languages define services by their profiles, but WSMO does not
explicitly represent the structure of composition of Web services in terms of
control flow.  Interactions of services are controlled by specific agents called
mediators which refer to the standard execution environment WSMX.

As can be noted, many of the latter proposals do not base their compositions on
a logic representation of pre- and post- conditions or on the function mapping
inputs to outputs.  Moreover, whenever a composition of sub services is defined,
this does not explicitly depend on the condition stated in declarative
description of the composite service.  A kind of composition that depends on the
formulas of service profiles can be found in the related approaches for action
formalisms and planning over description logics as
in~\cite{BaLuMiSaWo:05,CalDeGLenRos:07,DreThie:07,Milicic:07}. However,
classical planning techniques mostly generate sequences of services achieving a
goal: as noted in~\cite{TraversoP:04}, this can be limiting in practical cases
when one needs to distinguish between different condition cases and represent
sequence constraints between goals.

To compare our approach with the ones cited above, we remark that $\PAPdl$
assures that the composite service specification (the service profile) directly
follows from composition proof.  The correctness of compositions can be checked
directly by verifying the applicability conditions of the rules used in the
composition: moreover, our rules directly represent common control structures,
thus allowing to represent complex compositions.

Even if in this paper we 
detailed manual composition of services, by implementing $\PAPdl$ we would
obtain a method for automatic composition.  
For an actual implementation of our
calculus we need both an implementation of \BCDLZ and a method to convert in our
formalism the service descriptions from specification languages as OWL-S
that preserves their intended semantics.
The main limitation of our approach stands in the restricted expressivity of the
description logic at its base, that represents a small fragment of the actual
expressivity of current ontology languages.
In order to define a calculus for automatic composition,
in our future work we plan to study the properties of $\PAPdl$ and of
its underlying constructive description logic $\BCDLZ$. 
In particular, we remark that $\BCDLZ$ is related to the logic 
$\KALC$~\cite{BozFerFioFio:10} for which we already presented a decidable tableaux calculus:
moreover, it can be shown that the derived tableaux procedure is
{\sc Pspace}-complete, as in the case of satisfiability procedures for classical 
\ALC~\cite{SchSmo:91}. 
The relations between $\BCDLZ$, $\BCDL$ and $\KALC$ will be subject of our
future investigations.
On the other hand, in order to evaluate the properties of the composition calculus,
we also intend to examine its relations with software synthesis and action formalisms.

\bibliographystyle{plain}
\bibliography{BozFer10-CoRR}

\end{document}